\documentclass[
twocolumn,
]{ceurart}

\usepackage{listings}
\usepackage{tabularx}

\begin{document}

%%
%% Rights management information.
%% CC-BY is default license.
\copyrightyear{2023}
\copyrightclause{Copyright for this paper by its authors.
  Use permitted under Creative Commons License Attribution 4.0
  International (CC BY 4.0).}

%%
%% This command is for the conference information
\conference{8th Swiss Text Analytics Conference 2023, Neuchâtel, Switzerland, Jun. 12-14, 2023}

%%
%% The "title" command
\title{Improving Metrics for Speech Translation}

%%
%% The "author" command and its associated commands are used to define
%% the authors and their affiliations.

\author{Claudio Paonessa}[%
email=claudio.paonessa@fhnw.ch,
]
\cormark[1]
\fnmark[1]
\address{Institute for Data Science, FHNW University of Applied %%Sciences and Arts Northwestern Switzerland,
  Klosterzelgstrasse 2, 5210 Windisch, Switzerland}

\author{Dominik Frefel}[%
email=dominik.frefel@fhnw.ch
]
\fnmark[1]

\author{Manfred Vogel}[%
email=manfred.vogel@fhnw.ch
]

%% Footnotes
\cortext[1]{Corresponding author.}
\fntext[1]{These authors contributed equally.}

%%
%% The abstract is a short summary of the work to be presented in the
%% article.
\begin{abstract}
  We introduce Parallel Paraphrasing ($\text{Para}_\text{both}$), an augmentation method for translation metrics making use of automatic paraphrasing of both the reference and hypothesis. This method counteracts the typically misleading results of speech translation metrics such as WER, CER, and B{\scriptsize LEU} if only a single reference is available. We introduce two new datasets explicitly created to measure the quality of metrics intended to be applied to Swiss German speech-to-text systems. Based on these datasets, we show that we are able to significantly improve the correlation with human quality perception if our method is applied to commonly used metrics.
\end{abstract}

%%
%% Keywords. The author(s) should pick words that accurately describe
%% the work being presented. Separate the keywords with commas.

%%
%% This command processes the author and affiliation and title
%% information and builds the first part of the formatted document.
\maketitle

\section{Introduction}
For most practical use cases of speech-to-text systems, the transcription is perfect if the sentences are grammatically correct and the semantic meaning is fully captured. For speech translation, the phrasing of the transcription is ambiguous, contrary to word-by-word transcriptions for cases where input and output languages match. This means that a single reference rarely adequately covers the valid output range. The widely applied metrics B{\small LEU} \cite{papineni-etal-2002-bleu}, Word Error Rate (WER), and Character Error Rate (CER) fail to handle the occurring sentence ambiguity, if calculated based on single references per sample. Many additional references are needed for each sample to counteract this flaw, which are often not available due to the high cost and effort of collecting them. This leads to metric results often not agreeing with the human perception of the transcription quality. More generally, the ambiguous translation space and therefore the need for multiple references occurs for all translation tasks.

Swiss German is a family of dialects and a mostly spoken language. The language lacks standardized writing, the written form usually only appears in informal text messages. Therefore, Swiss speech-to-text systems transform Swiss German speech to Standard German text, i.e., speech translation with a very similar source and target language. Because Swiss German is different from Standard German regarding phonetics, vocabulary, morphology, and syntax, the system output tends not to match with the single reference in a test set. Table \ref{tab:exp} contains examples where common metrics fail to handle the paraphrased but semantically correct hypothesis.

\begin{table}[h]
\caption{Examples of semantically matching but paraphrased references and hypotheses with corresponding metric values. See Section \ref{sec:method} for details on the applied metrics.}
\label{tab:exp}
  \begin{tabularx}{\linewidth}{l l X}
    (1) & \textit{Ref:} & \textit{Gesucht wurde auch im nahen Ausland.} \\
     & Hypo: & Auch im nahen Ausland wurde gesucht. \\
     \cmidrule{2-3}
     & & B{\small LEU}: \textbf{0.562} | WER: \textbf{0.667} | CER: \textbf{0.771} \\ \\
    (2) & \textit{Ref:} & \textit{Der Spatenstich fand im Oktober letzten Jahres statt.} \\
     & Hypo: & Der Spatenstich fand letztes Jahr im Oktober statt. \\
     \cmidrule{2-3}
     & & B{\small LEU}: \textbf{0.271} | WER: \textbf{0.500} | CER: \textbf{0.404} \\ \\
    (3) & \textit{Ref:} & \textit{Überlegungen die Lage in Zukunft zu verbessern sind in Planung.} \\
     & Hypo: & Gedanken wie man die Lage zukünftig besser machen kann sind in Planung. \\
     \cmidrule{2-3}
     & & B{\small LEU}: \textbf{0.159} | WER: \textbf{0.700} | CER: \textbf{0.532} \\
  \end{tabularx}
\end{table}

%\begin{table}[h]
%  \begin{tabularx}{\linewidth}{l l X}
%    (4) & \textit{Ref:} & \textit{Diesem Witz mangelt es an Glaubwürdigkeit. (weitere Option)} \\
%     & Hypo: & Dem Witz fehlt es an Glaubwürdigkeit. \\
%     \cmidrule{2-3}
%     & & B{\small LEU}: \textbf{0.0} | WER: \textbf{0.0} | CER: \textbf{0.0} \\
%  \end{tabularx}
%  \refstepcounter{table}\label{exp3}
%\end{table}

Thanks to recent advancements in paraphrasing systems based on neural machine translation \cite{thompson-post-2020-paraphrase}, the quality of paraphrases significantly  increased. State-of-the-art paraphrasing systems not only cover synonym substitution but more advanced changes in sentence structures. Existing research improving translation metrics based on automatic paraphrasing \cite{bawden2020study} focuses on the augmentation of references and approaches to increase diversity to maximize coverage of the translation space.

Building  upon the existing idea of augmenting the references for an improved B{\small LEU} metric by automatic paraphrasing, we extend this method to also generate paraphrases of the hypothesis. Our experiments show that this addition significantly improves the correlation of common metrics with human perception.

\section{Related Work}

The use of paraphrasing for evaluating machine translation systems to address some of the weaknesses of popular metrics has a long history. The metrics METEOR \cite{banerjee-lavie-2005-meteor}, Meteor Universal \cite{denkowski-lavie-2014-meteor}, and ParaEval \cite{zhou-etal-2006-paraeval} support synonym matching, covering the simplest form of paraphrasing. A more recent approach ParaBLEU\cite{https://doi.org/10.48550/arxiv.2107.08251} includes a learned neural metric based on paraphrase representation learning, achieving state-of-the-art performance on the WMT Metrics Shared Task 2017 (WMT17) \cite{bojar-etal-2017-results}. \citeauthor{bawden2020study}, \citeyear{bawden2020study} \cite{bawden2020study} showed that slight gains to the correlation with human judgment can be expected with automatic paraphrasing to generate additional references for B{\small LEU}. In a more general analysis from \citeauthor{freitag-etal-2020-bleu} \citeyear{freitag-etal-2020-bleu} \cite{freitag-etal-2020-bleu} the impact of well-chosen references on correlation with human judgment for English to German translation is analyzed. The researchers found that a precisely defined paraphrasing task executed by professional linguists increases the correlation compared to backtranslation or other automated methods.

\section{Approach}

Compared to existing approaches that introduce automatic paraphrasing to the calculation of the B{\small LEU} metric \cite{kanayama-2003-paraphrasing, bawden2020study}, our approach is not limited to paraphrasing for the generation of diverse references but also generates paraphrases for the system output. Because the paraphrasing models based on machine translation typically have limited diversity of generated sentence structures, it is up to chance whether the sentence structure of a correct system output can be reproduced or not. With our approach of paraphrasing both the reference and the hypothesis ($\text{Para}_\text{both}$) we aim to increase the chances for an intended match and therefore limit the number of diverse paraphrases needed.

For the generation of paraphrases in German, we use a paraphrasing algorithm \cite{thompson-post-2020-paraphrase} based on the Prism translation model \cite{thompson-post-2020-automatic}. This algorithm pushes the output away from the input in the lexical space by penalizing n-gram overlaps. The algorithm penalizes n-gram (1-, 2-, 3-, and 4-grams) overlaps by subtracting values from the output log probabilities of the NMT model before selecting candidates during beam search. With exponential weighting on the penalization, the method ensures penalizing larger n-gram overlaps more harshly than smaller ones. A parameter $\alpha$ controls how much the model pushes the output away from the input during decoding. We use the paraphrasing algorithm and the translation model with the parameters from \citeauthor{thompson-post-2020-paraphrase} \citeyear{thompson-post-2020-paraphrase} \cite{thompson-post-2020-paraphrase} to sample the top $n$ backtranslation candidates. This corresponds to $\alpha=0.003$, $\beta=4$, and a beam width of $n$. Examples of resulting paraphrases are reported in Appendix \ref{app:examples}.

With $n$ additional generated paraphrases of the hypothesis and reference, we calculate $n+1$ metric values per sample for metrics supporting multiple references (e.g., B{\small LEU}). Single reference metrics (e.g., WER, CER) produce ${(n+1)}^2$ values. We explore different methods to aggregate the resulting values.

\section{Evaluation}

We consider two annotated data sources to measure the impact of our method if applied to established metrics. The two datasets Human Sentence Ratings (\textit{GER-HSR-1K}) and Online Transcription Ratings (\textit{GER-OTR-691}), specifically created for this work, are described in the following sections.

\subsection{Human Sentence Ratings}

We suggest a novel rating system aiming at capturing the essential components for comparing reference sentences and corresponding hypotheses for Standard German. The rating system consists of three binary values and one discrete rating with values between 0 and 3:

\begin{itemize}
  \item Hypothesis Grammar (case-insensitive and without punctuation)
  \item Hypothesis Punctuation
  \item Hypothesis Capitalization
  \item Semantic similarity rating
\end{itemize}

The first three binary values only indicate correctness or an error regarding grammar, punctuation, and capitalization. The fourth rating corresponds to a semantic comparison between the reference and the hypothesis as defined in Table \ref{tab:sem}.

\begin{table}
  \caption{Semantic similarity rating range.}
  \label{tab:sem}
  \begin{tabularx}{\linewidth}{l X}
    \toprule
    \textbf{Rating} & \textbf{Description} \\
    \midrule
    3 & Reference message is completely and unambiguously captured in the hypothesis (even possible if there are grammar or spelling mistakes)  \\
    \midrule
    2 & Virtually matching semantic meaning with only lack or abundance of insignificant details, misspelling of named entities is also considered an insignificant detail \\
    \midrule
    1 & Majority of reference message is captured and only a small significant part is semantically not matching with the reference \\
    \midrule
    0 & Majority of reference message is not captured \\
    \bottomrule
  \end{tabularx}
\end{table}

Based on this approach, we annotated a dataset with 1000 samples, denoted as \textit{GER-HSR-1K}. The samples originate from the two datasets Swiss Parliaments Corpus \cite{DBLP:conf/swisstext/PlussNSV21}, and SDS-200 \cite{pluss-etal-2022-sds}, containing Swiss German audio with Standard German transcriptions. The audio samples from these datasets were transcribed using a speech-to-text system, finetuned for Swiss German and based on the fairseq S2T \cite{ott2019fairseq, wang2020fairseqs2t} model. We replicate the Transformer baseline model architecture and training procedure from \citeauthor{pluss-etal-2022-sds}, \citeyear{pluss-etal-2022-sds} \cite{pluss-etal-2022-sds}. With this model, we generate realistic samples occurring in Swiss German to German speech translation. However, the references and corresponding hypotheses were only considered if they have a Levenshtein text distance greater than zero in order to exclude identical reference/hypothesis pairs.

\begin{figure}
  \centering
  \includegraphics[width=5.5cm]{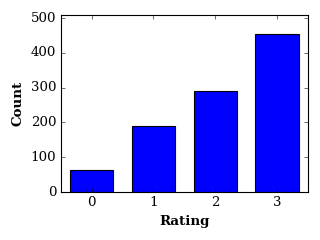}
  \caption{The distribution of the human-annotated semantic similarity rating in the \textit{GER-HSR-1K} dataset.}
  \label{fig:dist_hsr}
\end{figure}

In Figure \ref{fig:dist_hsr}, we report the distribution of the semantic similarity ratings. Because the sentences are sampled from a speech-to-text model achieving state-of-the-art results for Swiss German, lower ratings are less frequent and almost half of the samples are rated as perfectly matching the semantic meaning of the reference.

Our human annotators found a grammar, punctuation, or capitalization error in only 3.2\% of the \textit{GER-HSR-1K} hypotheses. The models mostly produce grammatically sound sentences. For the analysis in this work, we only make use of the semantic similarity rating to evaluate our metric.

The fully annotated dataset is available online.\footnote{https://www.cs.technik.fhnw.ch/i4ds-datasets}

\subsection{Online Transcription Ratings}

Through an online demo application showcasing a Swiss German speech-to-text system, we collected audio samples. After the transcription is shown to the user, they can voluntarily give feedback in the form of a discrete rating between 1 and 5 stars, with 1 star being the worst and 5 stars being the best rating. There is no further instruction given on how to rate the sentences. This dataset is extended by a human transcription to have a single reference ground-truth sentence.

Compared to the \textit{GER-HSR-1K} dataset, the sentences collected in this dataset are on average shorter and less representative of naturally spoken language. Due to the lack of detailed instructions, the crowd-sourced ratings tend to be inconsistent. Additionally, because the ground-truth transcription is loosely based on the system hypothesis, little paraphrasing occurs.

A filtered version of this dataset with 691 samples is denoted as \textit{GER-OTR-691}. This subset only contains pairs of references and hypotheses with a Levenshtein distance greater than zero.

\subsection{Evaluation Method}
\label{sec:method}

To estimate and compare the quality of different metrics, we calculate correlations (linear: Pearson's $r$, monotonic: Kendall's $\tau$) between metric results and human-annotated ratings. Specifically, we apply Kendall's Tau-like formulation defined in the WMT18 Metrics Shared Task \cite{ma-etal-2018-results}. The adaptation of the Kendall's Tau coefficient is defined as follows:

\begin{equation}
    \tau = \frac{|Concordant|-|Discordant|}{|Concordant|+|Discordant|} \\
\end{equation}

Whether a comparison between human judgment and a metric of a pair of distinct samples, $s_1$ and $s_2$, is counted as concordant (Conc) or discordant (Disc) is defined in the following matrix, where $m(s_i)$ and $h(s_i)$ are the metric value and the human rating of the sample $s_i$, respectively:

\begingroup
\setlength{\tabcolsep}{4pt}

\begin{table}[h]
  \begin{tabular}{l c | c c c}
       & & \multicolumn{3}{c}{Metric} \\
       & & \scalebox{0.7}{$m(s_1) < m(s_2)$} & \scalebox{0.7}{$m(s_1) = m(s_2)$} & \scalebox{0.7}{$m(s_1) > m(s_2)$} \\
      \specialrule{\cmidrulewidth}{0pt}{0pt}
      \multirow{3}{*}{\rotatebox{90}{Human}} & \scalebox{0.7}{$h(s_1) < h(s_2)$} & Conc & Disc & Disc \\
       & \scalebox{0.7}{$h(s_1) = h(s_2)$} & - & - & - \\
       & \scalebox{0.7}{$h(s_1) > h(s_2)$} & Disc & Disc & Conc \\
  \end{tabular}
\end{table}

\endgroup

This formulation means we exclude all human ties. In the case of non-identical human judgments, metric ties are always counted as $Discordant$. For this correlation coefficient to be consistent, we must ensure that human judgment and the metric have the same orientation, i.e., a higher score indicating higher transcription quality.

We use the sentence-level B{\small LEU} formula from sacreB{\small LEU} \cite{post-2018-call} with exponential smoothing. For the WER and CER metrics we employ the corresponding accuracy rate ($1 - \textit{Error-rate}$) to properly align the scores for Kendall's Tau-like formulation. We normalize the text by removing all punctuation and transforming all characters to lowercase.

\section{Results}
We analyzed different aggregation methods: maximum, minimum, average, or top-n-averaging. We found the minimum and the average to decrease the correlation with human judgment compared to the maximum. In some cases, averaging a subset of the top values outperformed the maximum, but the results were inconsistent. Because ties between metric values of two distinct samples are discouraged in the used Kendall's Tau-like formulation, the maximum, and an additional arbitrarily small random number added outperforms all other methods by decreasing the chances of ties to virtually zero. Statistically, in 50\% of the cases where a tie would occur, we randomly hit the correct ranking. For a fair comparison, we report results without adding this random value. The added random number only helps to artificially increase Kendall's Tau-like values. It does not improve the metric in any useful way. All the results are reported with the maximum as the aggregation function.

\begin{table}[htb]
  \caption{Kendall's $\tau$ and Pearson's $r$ correlations on the \textit{GER-HSR-1K} (HSR) and the \textit{GER-OTR-691} (OTR) datasets. $\text{Para}_\text{ref}$ ($n$) refers to the multi-reference version with $n$ additional generated references. Metrics denoted as $\text{Para}_\text{both}$ ($n$) refer to our suggested extended version with the hypothesis and references augmented with $n$ additional generated paraphrases.}
  \label{tab:res}
  \begin{tabularx}{\linewidth}{l X c c}
    \toprule
    \textbf{Dataset} & \textbf{Metric} & \textbf{$\tau$} & \textbf{$r$} \\
    \midrule
    \multirow{9}{*}{\textbf{HSR}} & WER & 0.3472 & 0.4704 \\
    & $\text{Para}_\text{ref}$ \footnotesize{(11)} WER & 0.3725 & 0.4879 \\
    & $\text{Para}_\text{both}$ \footnotesize{(6)} WER & \textbf{0.5115} & \textbf{0.6137}  \\
    \cmidrule{2-4}
    & CER & 0.2632 & 0.3511 \\
    & $\text{Para}_\text{ref}$ \footnotesize{(11)} CER & 0.3081 & 0.3982  \\
    & $\text{Para}_\text{both}$ \footnotesize{(6)} CER & \textbf{0.4811} & \textbf{0.5513} \\
    \cmidrule{2-4}
    & B{\footnotesize LEU} & 0.3167 & 0.3903 \\
    & $\text{Para}_\text{ref}$ \footnotesize{(11)} B{\footnotesize LEU} & 0.3798 & 0.4438 \\
    & $\text{Para}_\text{both}$ \footnotesize{(6)} B{\footnotesize LEU} & \textbf{0.4892} & \textbf{0.5872} \\
    \midrule
    \multirow{9}{*}{\textbf{OTR}} & WER & 0.5903 & 0.6133  \\
    & $\text{Para}_\text{ref}$ \footnotesize{(8)} WER & 0.5957 & 0.6191  \\
    & $\text{Para}_\text{both}$ \footnotesize{(2)} WER & \textbf{0.5972} & \textbf{0.6238}  \\
    \cmidrule{2-4}
    & CER & 0.6283 & 0.6516 \\
    & $\text{Para}_\text{ref}$ \footnotesize{(8)} CER & 0.6356 & 0.6590  \\
    & $\text{Para}_\text{both}$ \footnotesize{(2)} CER & \textbf{0.6369} & \textbf{0.6632}  \\
    \cmidrule{2-4}
    & B{\footnotesize LEU} & 0.5531 & 0.6419 \\
    & $\text{Para}_\text{ref}$ \footnotesize{(7)} B{\footnotesize LEU} & 0.5369 & 0.6270 \\
    & $\text{Para}_\text{both}$ \footnotesize{(2)} B{\footnotesize LEU} & \textbf{0.5578} & \textbf{0.6473} \\
    \bottomrule
  \end{tabularx}
\end{table}

The results for the two reference datasets are reported in Table \ref{tab:res}. For both augmentation methods, we report results for the best number of paraphrases between 1 and 16. We show a significant improvement of the correlations for the \textit{GER-HSR-1K} dataset if applying $\text{Para}_\text{both}$. With the augmentation limited to the references ($\text{Para}_\text{ref}$), we see a much lower improvement to the baseline. On the \textit{GER-OTR-691} dataset we observe very limited gains. Because of the lack of paraphrasing occurring in this dataset, we did not expect a large improvement in the correlations.

\begin{figure}[h]
  \centering
  \includegraphics[width=\linewidth]{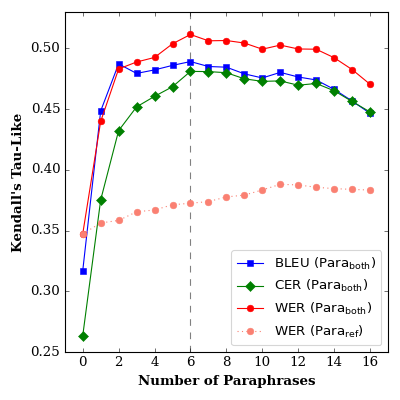}
  \caption{Results of the $\text{Para}_\text{both}$ method applied to three metrics on the \textit{GER-HSR-1K} dataset with different numbers of paraphrases (for both the reference and the hypothesis). Additionally, the WER results with only the references augmented ($\text{Para}_\text{ref}$). Zero paraphrases correspond to no augmentation and the single-reference version of the metric.}
  \label{fig:numb_para}
\end{figure}

In Figure \ref{fig:numb_para}, we show the impact of the number of paraphrases on the correlation with human judgment. Throughout all metrics six seems to be the best number of paraphrases for our method $\text{Para}_\text{both}$. The parameterization of the underlying automatic paraphrasing model and the dataset impacts the number of paraphrases. Therefore, the chosen number of paraphrases should not be interpreted as a generally good pick. Additionally, we include the result of the WER metric with the augmentation limited to references ($\text{Para}_\text{ref}$). In this case, the strongest correlation between human ratings and automatic metrics is achieved at 11 paraphrases, but is significantly lower than our parallel approach.

\begin{figure}[h]
  \centering
  \includegraphics[width=\linewidth]{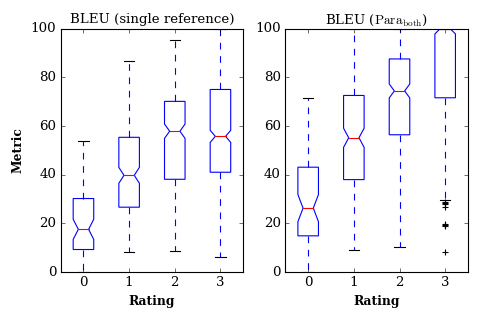}
  \caption{A comparison between the distributions of the single reference B{\small LEU} and the $\text{Para}_\text{both}(6)$ B{\small LEU} score for the different semantic similarity ratings in the \textit{GER-HSR-1K} dataset.}
  \label{fig:comp}
\end{figure}

In addition to the numeric correlation, we show a visual comparison of human ratings and metric values in Figure \ref{fig:comp}. Based on the single reference B{\small LEU} scores, we observe that the metric clearly underestimates a lot of samples with a high human rating of 2 or 3. The metric distributions of the samples with these high ratings are nearly indistinguishable. Our approach improves these distributions and better aligns the metrics to be linearly increasing with the human ratings. Because our approach combined with maximum aggregation can only increase the metric value, we overestimate some of the low-rated samples.

\section{Conclusion}

In this paper, we introduced $\text{Para}_\text{both}$, an augmentation method for translation metrics. We demonstrated a significant increase in correlation with human judgment for Swiss German to German speech translation. Based on  state-of-the-art automatic paraphrasing, we produce different paraphrased versions of the reference and the hypothesis. Using our method, we improve the robustness of existing metrics by addressing paraphrasing that may arise in translation tasks.

Based on encouraging side experiments conducted on the WMT19 \cite{fonseca-etal-2019-findings} dataset, we propose further investigation into the overall effectiveness of this approach in neural machine translation. We observed a relative increase of 52\% on Kendall's Tau-like score with $\text{Para}_\text{both}$.

For future work, we also recommend exploring methods to mitigate the overestimation of low-rated samples. Our current method does not include any measures to reduce the risk of overestimation. Additionally, exploring other paraphrasing methods and their parameter space may also lead to more suitable paraphrases.

The novel dataset \textit{GER-HSR-1K} is made publicly available to help advance the development of more appropriate metrics for speech translation, especially Swiss German.

\section{Acknowledgement}
This work was supported by HASLER Foundation within the project ”Quality metrics for Swiss German speech recognition” [22031].

%%
%% The acknowledgments section is defined using the "acknowledgments" environment
%% (and NOT an unnumbered section). This ensures the proper
%% identification of the section in the article metadata, and the
%% consistent spelling of the heading.
%%\begin{acknowledgments}
%%  This work was supported by funding from the Hasler Foundation under the grant agreement No %%TODO.
%%\end{acknowledgments}

%%
%% Define the bibliography file to be used
\bibliography{cvww23-bib}

\clearpage

%%
%% If your work has an appendix, this is the place to put it.
\appendix

\section{Paraphrase Examples}
\label{app:examples}

\noindent
\begin{tabularx}{\textwidth}{l X}
\cmidrule{1-2}
\textit{Ref:} & \textit{Gesucht wurde auch im nahen Ausland.} \\
 \cmidrule{1-2}
 & Es wurde auch im nahen Ausland gesucht. \\
 & Es wurde auch im benachbarten Ausland gesucht. \\
 & Auch im benachbarten Ausland wurde gesucht. \\
 & Auch im benachbarten Ausland suchte man. \\
 & Es wurde auch im nahen Ausland verfolgt. \\
 & Es wurde auch im benachbarten Ausland verfolgt. \\
 \cmidrule{1-2}
 \textit{Hypo:} & \textit{Auch im nahen Ausland wurde gesucht.} \\
 \cmidrule{1-2}
 & Ebenfalls im benachbarten Ausland wurde gesucht. \\
 & Auch im benachbarten Ausland wurde gesucht. \\
 & Ebenfalls im benachbarten Ausland suchte man nach. \\
 & Ebenfalls im benachbarten Ausland suchte man. \\
 & Auch im benachbarten Ausland suchte man. \\
 & Ebenfalls im nahen Ausland wurde gesucht. \\
\end{tabularx}

\vspace{1cm}

\noindent
\begin{tabularx}{\textwidth}{l X}
\cmidrule{1-2}
\textit{Ref:} & \textit{Der Spatenstich fand im Oktober letzten Jahres statt.} \\
 \cmidrule{1-2}
&Der Spatenstich geschah im Oktober letzten Jahres. \\
&Der Spatenstich geschah im Oktober vergangenen Jahres. \\
&Der Spatenstich geschah im vergangenen Oktober. \\
&Der Spatenstich fand letztes Jahr im Oktober statt. \\
&Es geschah im Oktober letzten Jahres. \\
&Es geschah im Oktober vergangenen Jahres. \\
\cmidrule{1-2}
\textit{Hypo:} & \textit{Der Spatenstich fand letztes Jahr im Oktober statt.} \\
 \cmidrule{1-2}
&Der Spatenstich fand im letzten Jahr im Oktober statt. \\
&Der Spatenstich fand im vergangenen Jahr im Oktober statt. \\
&Der Spatenstich fand im letzten Jahr im vergangenen Oktober statt. \\
&Der Spatenstich fand im Oktober des vergangenen Jahres statt. \\
&Der Spatenstich fand im letzten Jahr im Oktober ab. \\
&Es fand im letzten Jahr im Oktober statt. \\
\end{tabularx}

\vspace{1cm}

\noindent
\begin{tabularx}{\textwidth}{l X}
\cmidrule{1-2}
\textit{Ref:} & \textit{Überlegungen die Lage in Zukunft zu verbessern sind in Planung.} \\
 \cmidrule{1-2}
&Überlegungen zur Verbesserung der Zukunftssituation sind geplant. \\
&Überlegungen zur Verbesserung der Situation in der Zukunft sind geplant. \\
&Überlegungen zur Verbesserung der Situation in der Zukunft sind in Planung. \\
&Erwägungen zur Besserung der zukünftigen Lage gibt es. \\
&Erwägungen zur Besserung der zukünftigen Lage sind geplant. \\
&Überlegungen wie zukünftig die Dinge besser gestaltet werden können, sind in Planung. \\
\cmidrule{1-2}
\textit{Hypo:} & \textit{Gedanken wie man die Lage zukünftig besser machen kann sind in Planung.} \\
 \cmidrule{1-2}
&Gedanken darüber, wie zukünftig die Dinge besser gestaltet werden können, sind in Planung. \\
&Gedanken darüber, wie zukünftig die Dinge besser gestaltet werden können, befinden sich in Planung. \\
&Ideen, wie sich die Situation in der Zukunft verbessern ließe, befinden sich in Planung. \\
&Ideen, wie sich die Situation in der Zukunft verbessern ließe, entwickeln sich. \\
&Jetzt geht es darum, darüber nachzudenken, wie man die Lage zukünftig besser machen kann. \\
&Überlegungen zur Verbesserung der Situation in der Zukunft sind im Gange. \\
\end{tabularx}
\end{document}